\documentclass[letterpaper, 10 pt, conference]{ieeeconf}  

\IEEEoverridecommandlockouts                              

\usepackage{cite}
\usepackage{ulem} 
\usepackage{balance}
\usepackage{graphicx} 
\usepackage{stfloats}
\usepackage{hyperref}   
\usepackage{float}
\usepackage[ruled]{algorithm2e}
\usepackage{caption}
\usepackage{subcaption}
\usepackage{multirow} 
\usepackage{booktabs}
\usepackage{indentfirst}
\usepackage{color}  
\usepackage{lipsum} 
\usepackage{amsmath,esint} 
\usepackage{cuted}
\usepackage{CJKutf8}
\usepackage{verbatim}
\usepackage{amssymb}
\usepackage{makecell}
\usepackage{bm}
\usepackage{setspace}
\usepackage{array,caption}
\usepackage{fancyhdr}
\usepackage[labelfont=bf]{caption}
\usepackage{ulem}
\usepackage{color}

\hypersetup{
	colorlinks=true,
	linkcolor=blue,
	citecolor=blue,
	urlcolor=blue,
}
\IEEEoverridecommandlockouts                              
\title{\LARGE \bf
Learning Quadrupedal Robot Locomotion for Narrow Pipe Inspection} 

\author{Jing Guo, Ziwei Wang, and Weibang Bai$^{*}$%
\thanks{
This work is supported by the Shanghai Pujiang Program (23PJ1408500), Shanghai Frontiers Science Center of Human-centered Artificial Intelligence (ShangHAI), MoE Key Laboratory of Intelligent Perception and Human-Machine Collaboration (KLIP-HuMaCo).
The experiments of this work were supported by the Core Facility Platform of Computer Science and Communication, SIST, ShanghaiTech University.
\textit{(Corresponding author: Weibang Bai)}
}
\thanks{Jing Guo and Weibang Bai are with the ShanghaiTech Automation and Robotics (STAR) Center, School of Information Science and Technology, ShanghaiTech University, Shanghai, 201210, China.  Ziwei Wang is with School of Engineering, Lancaster University, Lancaster, UK.}
}

\begin{document}

\maketitle
\thispagestyle{empty}
\pagestyle{empty}

\begin{abstract}
Various pipes are extensively used in both industrial settings and daily life, but the pipe inspection especially those with narrow sizes are still very challenging with tremendous time and manufacturing consumed. Quadrupedal robots, inspired from patrol dogs, can be a substitution of traditional solutions but always suffer from navigation and locomotion difficulties. In this paper, 
we introduce a Reinforcement Learning (RL) based method to train a policy enabling the quadrupedal robots to cross narrow pipes adaptively.
A new privileged visual information and a new reward function are defined to tackle the problems. Experiments on both simulation and real world scenarios were completed, demonstrated that 
the proposed method can achieve the pipe-crossing task even with unexpected obstacles inside.
\end{abstract}

\section{INTRODUCTION}
\vspace{-1mm}
In daily life, patrol dogs are frequently deployed in search and rescue missions across complex environments, such as chaotic post-disaster sites, narrow passageways, and confined or extended pipelines. As quadruped animals, they exhibit a diverse range of running gaits, excelling in speed, stability, efficiency, and adaptability \cite{fan2024review}.
Likewise, bionic quadrupedal robots are increasingly utilized in similar circumstances.

Pipelines, particularly narrow and confined ones, present significant challenges in inspection and monitoring tasks. These environments are often characterized by limited space, complex geometries, and a lack of natural light and guidance information, making it difficult to navigate and inspect using conventional methods \cite{10033270}. The narrowness of these pipelines with unexpected obstacles restricts the movement of traditional inspection tools and robotic systems, often requiring specialized equipment to maneuver through tight bends, junctions, and varying diameters\cite{9625928}. Additionally, the confined space limits signal communication and visual feedback, making remote control and real-time decision-making more challenging. These challenges are exacerbated in scenarios involving hazardous or unknown conditions, such as gas leaks or structural weaknesses, where ensuring the safety and reliability of inspection becomes crucial\cite{chen2023decade,mills2017advances}.

One of the primary methods for addressing the challenges of narrow pipeline inspection task is the deployment of robotic inspection systems. These systems are specifically engineered to navigate confined spaces, featuring designs such as wheeled type, tracked type, wall-pressed type, walking type, inchworm type and screw type robots that can move through tight bends and complex geometries\cite{9731152,jang2022review,ambati2020review}. Equipped with various sensors, such as cameras, ultrasonic sensors, and laser scanners, these robots can detect structural defects like cracks, corrosion, and leaks in real-time, significantly enhancing inspection efficiency and accuracy\cite{Han2016/05,archila2013study,1391010}. Some advanced models even incorporate remote control \cite{wang2021multiple,QIU201693} or semi-autonomous navigation capabilities, allowing operators to manage inspections from a safe distance\cite{ismail2012development}.

However, traditional robotic inspection systems are often highly specialized, which limits their adaptability in diverse real-world applications. This will raise development and manufacturing costs and increase operational complexity, making training and operation less user-friendly and less efficient for navigation and inspection tasks\cite{mills2017advances,jang2022review}.
In contrast, bionic quadrupedal robots provide superior flexibility and adaptability thanks to their dynamic, legged locomotion, enabling them to maneuver more effectively in irregular environments, traverse obstacles, and maintain stability on uneven or slippery surfaces commonly encountered in real-world pipeline scenarios\cite{9636738,sun2020real}. 
The integration of real-time data processing and advanced diagnostic tools further enhances the robots' ability to detect and analyze pipeline anomalies on the spot, providing valuable insights for maintenance\cite{9959471}. These features make quadrupedal robots a highly promising alternative for pipeline inspection, as they combine automation, adaptability, and advanced control to achieve better performances.

The control of quadrupedal robots traditionally relies on classical model-based control methods such as MPC \cite{pandala2022robust}. These approaches are built upon accurate pre-defined models of the robot's mechanics and environment, which can be highly effective in structured and predictable environments \cite{bellicoso2017dynamic,liao2023walking}. 

However, they can also be computationally expensive and need to struggle with high-dimensional state spaces and non-linear dynamics, 
making difficult for the robot to find optimal solutions and adapt to unexpected or uncertain situations, such as slippery surfaces, complex terrains, or sudden external disturbances \cite{fang2023actuation}.

Recently, Reinforcement Learning (RL) based control is widely used to overcome the traditional limitations. RL allows quadrupedal robots to learn optimal control policies through interaction with the environment, without relying on pre-defined models. By continuously learning from trial and error, RL-based controllers can adapt to a wide range of scenarios, including complex terrains, dynamic obstacles, and varying environmental conditions\cite{bellegarda2022robust,wang2024learning}. 
Hoeller et al. \cite{hoeller2024anymal} introduces a fully learned control approach for quadrupedal robots to perform agile navigation in parkour-like environments using advanced locomotion skills trained in simulation. Kumar et al. \cite{kumar2021rma} presents the RMA algorithm, which enables real-time adaptation of quadrupedal robots to varying terrains and conditions using a two-part system: a base policy and an adaptation module, trained entirely in simulation without real-world fine-tuning. Cheng et al. \cite{cheng2024extreme} demonstrates a low-cost quadrupedal robot performing extreme parkour tasks such as high jumps, long jumps, and handstands using a single neural network trained end-to-end with reinforcement learning on visual input from a depth camera. Zhuang et al. \cite{zhuang2023robot} introduce a two-stage reinforcement learning method to train firstly with soft dynamics and then to fine-tune with hard dynamics to develop robust and agile locomotion, enabling the robots to handle various complex terrains without relying on pre-defined trajectories or explicit mapping.

Whereas, those RL approaches are not suitable for navigating through pipelines due to several inherent challenges. Firstly, pipelines are characterized by narrow and confined spaces that severely limit the robot's ability to perform complex maneuvers which may lead to frequent collisions and instability. Secondly, many of these methods rely on privileged information during training, such as precise terrain data, elevation maps, or height information, which are difficult to fully represent within a pipeline. Additionally, the reward functions used in these control strategies are often tailored for specific open-environment tasks, such as traversing obstacles or maintaining balance on varied terrain. However, these reward designs do not align well with the requirements of pipeline navigation, where the primary challenges include maintaining stability in confined spaces and avoiding contact with the side walls. 

In this regard, we propose a novel RL framework for quadrupedal robots navigating through narrow pipelines. The main contributions include:
\begin{itemize}
    \item Design a pipeline terrain which can be used in simulated RL training.
    \item Propose bidirectional Scandots which gives privileged visual information for  training in pipeline terrain.
    \item Design a simple reward function which is suitable for quadrupedal robot navigating through narrow pipelines.
    \item As the first RL-based solution, we develop and deply a policy trained by RL that can realize the locomotion of quadrupedal robot for navigating through narrow pipelines with various multi-modal onboard sensory information.
\end{itemize}

\vspace{-2mm}
\section{METHOD}
\begin{figure}[t]
    \centering
    \includegraphics[width=1\linewidth]{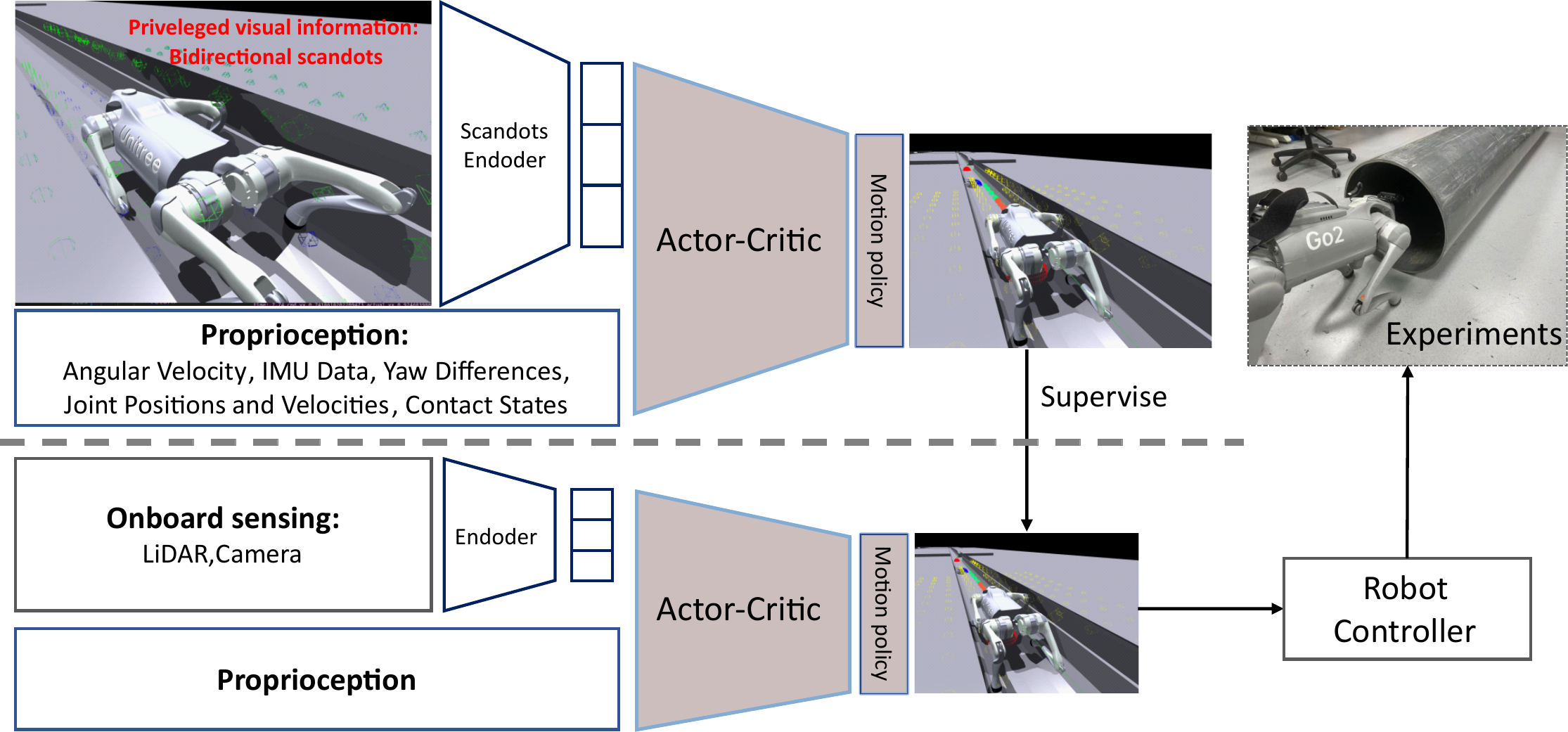}
    \caption{Overall framework of our method. In the first stage, We introduce bidirectional scandots and the proprioception as  the privileged information. In the second stage, we distill into a policy with depth camera as the input.}
    \label{fig:framework}
    \vspace{-6mm}
\end{figure}

In this study, to fulfill the narrow pipe inspection tasks robustly, we developed a reinforcement learning framework to train a quadrupedal robot to navigate through a pipe-like terrain, which includes complex and irregular features. The whole framework is shown in Fig.~\ref{fig:framework}. The core of our approach involves designing a realistic simulated environment, leveraging privileged information to enhance situational awareness, and carefully crafting reward functions to encourage desired behaviors. The following subsections detail the specific components of our methodology: terrain setup, privileged information design, and reward function construction.

\subsection{Terrain Setup}

To simulate the challenging environment of navigating through a pipe, we developed a custom terrain using a height field representation \cite{rudin2022learning}. This setup is designed to replicate realistic and variable conditions, essential for robust training of the robot. The terrain construction process involves generating a cylindrical pipe structure and introducing randomly added obstacles to enhance environmental complexity.

\subsubsection{Terrain Definition}
\begin{figure}[t]
    \centering
    \includegraphics[width=0.95\linewidth]{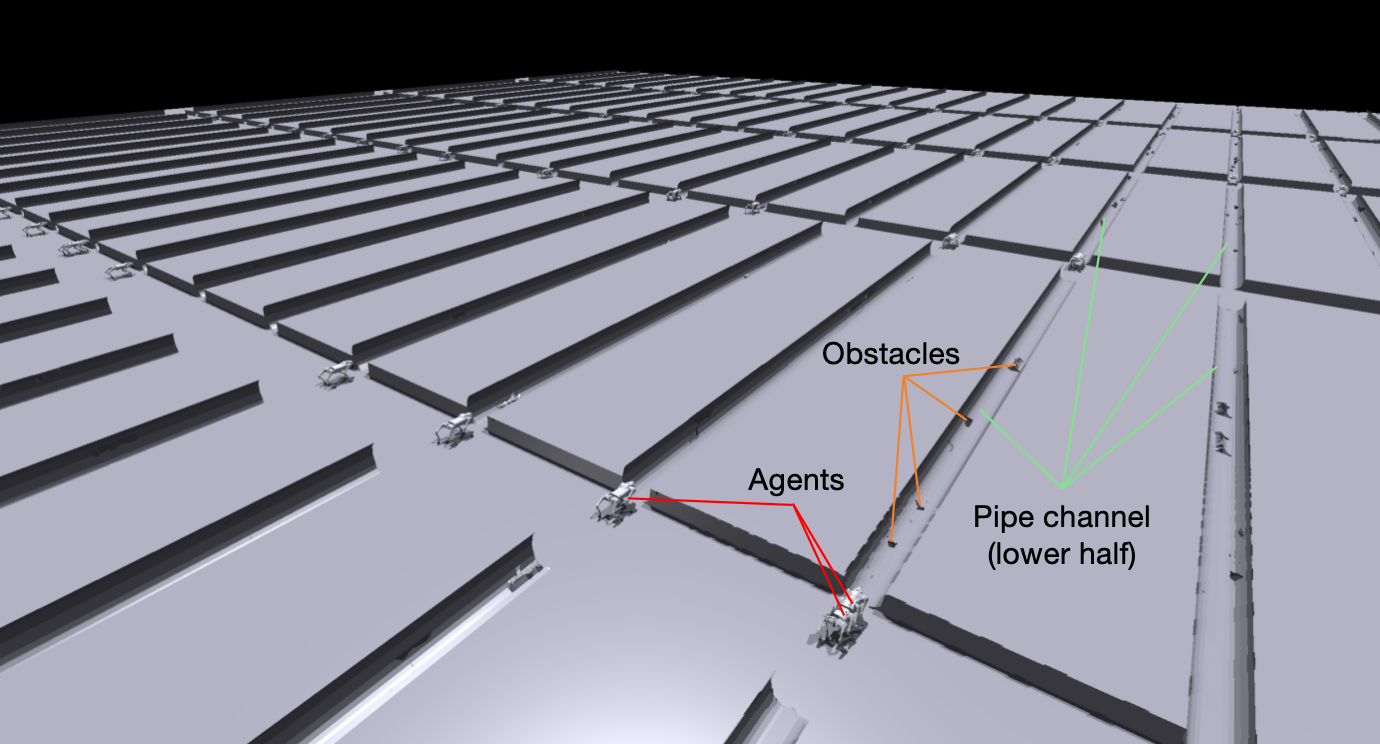}
    \caption{During training, the terrain with multiple pipe channels are designed, which consists of 10 rows and 40 columns. The bottom half of the pipe channels is exposed, for the better visualization and observation of the poses and states of the quadrupedal robots inside the pipe.}
    \label{fig:terrain_design}
    \vspace{-6mm}
\end{figure}

We define the whole terrain for training as a combination of many sub-terrains, as shown in Fig.~\ref{fig:terrain_design}. The rows of sub-terrain represents the level of training curriculum.
\begin{itemize}
    \item \textbf{Sub-terrain:} Each sub-terrain is defined as a rectangle of 18m*4m. The platform is 2m*4m while the other part of the sub-terrain is the pipe. 
    The robot will be deployed and begin its exploration from the center of the platform.

    \item \textbf{Training curriculum:} To enhance training efficiency and progressively improve the robot’s performance in complex terrains, we implemented a game-inspired curriculum learning strategy. We set a threshold: $\tau=v_x*T$, where $v_x$ is the linear velocity along $x$ axis while $T$ is the length of an episode. The agent who goes farther than the 0.8*threshold will level up, otherwise, who goes shorter than 0.4*threshold will level down. With the difficulty level increasing, the radius of pipe will decrease, which will lead to a harder task for the robot to cross. 
\end{itemize}

\subsubsection{Pipe Definition}

The primary terrain structure is modeled as a cylindrical height field, which creates a confined environment that the robot must navigate through. We create the pipe terrain with two stage, in the first stage, we use the pipe's key parameter, radius, to define a pipe to ensure a realistic and challenging navigation task. The height field is defined as follow:
\begin{equation}
    z(y)=\sqrt{r^2-(y-y_0)^2}+z_0 
\end{equation}
where $y_0$ is the y coordinate of the center line of the pipe, while $z_0$ is the height of the center line. Since the characteristic of the pipe, which is actually a cylinder whose center line parallel to the $x$ axis. So the function defines the height field is unrelated to $x$

\subsubsection{Obstacles}

To introduce variability and complexity, obstacles are randomly added within the pipe. These obstacles represent potential obstacles in the real pipe that the robot must detect and navigate around, increasing the difficulty of the task. The design details are:

     Randomly choose 0 to 4 obstacles and add them to the inner surface pipe. Each obstacle varies in height between 0.1 m and 0.3 m, with lengths ranging from 0.2 m to 0.5 m and width of 0.1m, ensuring diverse obstacle configurations.

\subsubsection{Training Techniques}

To bridge the gap between simulation and real-world deployment, we introduced several modifications in the training environment. These adjustments aimed to make the simulation more representative of the real-world scenarios the robot would encounter inside a pipe. The key modifications include:

\begin{itemize}
    \item \textbf{Reduced Pipe Wall Friction:} In real-world scenarios, the inner surface of pipes is often coated with substances like water, oil, or sand, which significantly reduce friction. To help the robot to learn to navigate under low-friction conditions, we set the friction like following: $f_\text{static}=0.2,~ f_\text{dynamic}=0.1$.

    \item \textbf{Raised Pipe Entrance:} The pipe entrance in the simulation was elevated slightly to mimic the real-world setup where the robot needs to transition smoothly into the pipe: $
    h_\text{entrance}=0.1 \text{m}$.
    
    \item \textbf{Random External Force Disturbance:} To prepare the robot for unexpected conditions inside the pipe,  a random external force disturbance was applied to the robot's base during training. 

\end{itemize}

\subsection{Observation Design}
\subsubsection{Problem Definition}
Our objective is to enable a quadruped robot to traverse through a challenging pipe environment with varying terrains and obstacles. This task is modeled as a Markov Decision Process (MDP) defined by $\mathcal{M} = (\mathcal{S}, \mathcal{A}, \mathcal{P}, \mathcal{R}, \gamma)$, where $\mathcal{S}$ is the state space, $\mathcal{A}$ is the action space, $\mathcal{P}$ denotes the state transition probabilities, $\mathcal{R}$ is the reward function, and $\gamma$ is the discount factor. The robot observes its state $s_t$ at each time step, takes an action $a_t$, and receives a reward $r_t$. The action space consists of joint commands for locomotion, and the reward function is designed to guide the robot towards stable and efficient traversal of the pipe environment.

The observation vector $\mathbf{o}_t$ at each time step $t$ provides comprehensive information about the robot's state and environment, facilitating robust decision-making during locomotion. The observation $\mathbf{o}_t \in \mathbb{R}^{m}$ is defined as:
\begin{equation}
    \mathbf{o}_t = [\mathbf{\omega}_t, \text{IMU}_t, 
    \mathbf{g},
    \mathbf{c}_t,
    \mathbf{q}_t, \mathbf{\dot{q}}_t, 
    \mathbf{a}_{t-1},
    \mathbf{\xi}_t]\in\mathbb{R}^{49}
\end{equation}
where the specific elements can be defined in Table~\ref{tab_elem_1}.
\begin{table}[h]
    \centering
    \caption{Elements of the proprioception}
    \setlength{\tabcolsep}{0.8mm}{
    \begin{tabular}{cl}
        \toprule
        Element & \makecell[c]{Meaning} \\ 
        \midrule
        $\mathbf{\omega}_{t}\in\mathbb{R}^{3}$ & Base angular velocity  \\
        $\text{IMU}_t\in\mathbb{R}^{2}$ & \makecell[l]{Roll and pitch angles from the inertial \\measurement unit}  \\
        $\mathbf{g}\in\mathbb{R}^{3}$ & The gravity parameter  \\
        $\mathbf{c}_t\in\mathbb{R}^{3}$ & \makecell[l]{Commanded velocity along the forward \\direction}  \\
        $\mathbf{q}_t\in\mathbb{R}^{12}$ & Joint positions, scaled by their limits \\
        $\mathbf{\dot{q}}_t\in\mathbb{R}^{12}$ & Joint velocities, scaled  \\
        $\mathbf{a}_{t-1}\in\mathbb{R}^{12}$ & Last applied action.  \\
        $\mathbf{\xi}_t\in\mathbb{R}^{296}$ & Privileged information vector  \\
        \bottomrule 
    \end{tabular}}
    \label{tab_elem_1}
\end{table}

The privileged information vector $\mathbf{e}$ is given by:
\begin{equation}
    \mathbf{\xi}_{t} = [\mathbf{h}_{\text{downwards}},\mathbf{h}_{\text{upwards}},\mathbf{v}_{t},\mathbf{e}_{t}]\in\mathbb{R}^{296}
\end{equation}
where $\mathbf{h}_{\text{downwards}}\in\mathbb{R}^{132}$ is the downwards scanning heights given by 11*12 sample points, $\mathbf{h}_{\text{upwards}}\in\mathbb{R}^{132}$ is the upwards scanning heights given by 11*12 sample points, and $\mathbf{v}_t \in\mathbb{R}^{3}$ is the base linear velocity. $\mathbf{e}_{t}\in\mathbb{R}^{29}$ is the environment vector consists of a mass parameter $m_{t}\in\mathbb{R}^4$, a friction parameter $\mu\in\mathbb{R}^1$, and motor strength $s_{t}\in\mathbb{R}^{24}$.

\subsubsection{Privileged Information Design} 
\begin{figure}[h]
\vspace{-1mm}
    \centering
    \includegraphics[width=1\linewidth]{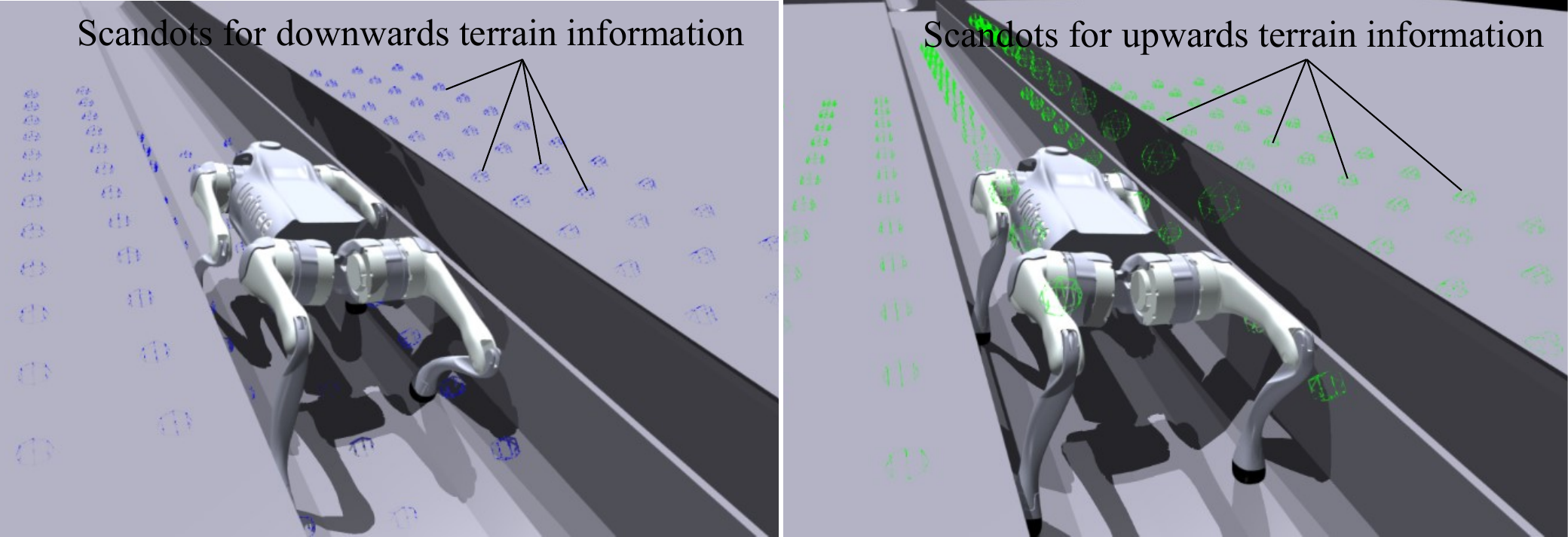}
    \caption{Design of bidirectional scandots for obtaining both the ceiling (top half) and the floor (bottom half) terrain information of the pipe.}
    \label{fig:3}
    \vspace{-5mm}
\end{figure}

For providing the agent with enhanced environmental information which contains both upwards and downwards terrain information, we introduce a method called the bidirectional height scanning, which is used as the privileged visual information during the first RL training stage. This privileged visual information will be encoded to a latent vector represents the environment in the pipe:
\begin{itemize}
    \item \textbf{Bidirectional Height Scanning:} We implemented an 11×12 grid of sampling points around the robot, ensuring consistent spacing to achieve uniform coverage across its body which is, as shown in Fig.~\ref{fig:3}. Additionally, a supplementary upward-facing grid captures height data from the pipe's ceiling and walls. For the upwards scanning data, we collect the measured information with the same point matrix to ones downwards. We give a virtual ceiling to the terrain and the scanning data will collect the distance from the robot base to the ceiling.
    \item \textbf{Privileged Data Encoding:} This rich set of height information is processed using a specialized encoder network, such as a feedforward neural network, which encodes the data into a compact form that is fed into the policy network.
\end{itemize}

\subsection{Reward Function Design}

The reward function design is crucial to guiding the robot's behavior as it navigates the complex pipe environment. To ensure effective learning, we incorporated a multi-faceted reward strategy focusing on velocity tracking, energy efficiency, collision avoidance, and maintaining a central position in the pipe. The main reward components are detailed below:

\subsubsection{Velocity Tracking}
Velocity tracking rewards are designed to encourage the robot to follow the desired velocity commands closely. The reward for tracking the velocity consists of tracking linear velocity and angular velocity. Tracking linear velocity reward introduces the heading direction \cite{cheng2024extreme}:
\begin{align}
& r_{\text{lin\_vel}} =\lambda_{\text{lin\_vel}}\frac{\min(\mathbf{v}_{\text{target}} \cdot \mathbf{v}_{\text{current}}, c_{\text{vel}})}{c_{\text{vel}} + C}, \\
& r_{\text{ang\_vel}} = \lambda_{\text{ang\_vel}}e^{-|\mathbf{\omega}_{\text{yaw}}^{\text{cmd}}-\mathbf{\omega}_{\text{yaw}}|}.
\end{align}
where $\mathbf{v}_{\text{target}}$ is the target velocity, $\mathbf{v}_{\text{current}}$ is the robot's current velocity, and $c_{\text{vel}}$ is the command velocity magnitude. $C$ is a common value here. We give $C=1*10^{-5}$ in training. While $\mathbf{\omega}_{\text{yaw}}^{\text{cmd}}$ is the target angular velocity, $\mathbf{\omega}_{\text{yaw}}$ is the robot'current velocity.

\subsubsection{Energy Consumption}

To promote energy-efficient behavior, penalties are applied based on the robot's torque usage and joint accelerations. This helps minimize unnecessary movements and encourages smoother actions. The energy-related reward functions are given by\cite{rudin2022learning}:
\begin{align}
r_{\text{torque}} &= -\lambda_{\text{torque}} \sum_{i=1}^{n} \tau_i^2, \\
r_{\text{delta\_torques}} &= -\lambda_{\text{delta\_torques}} \sum_{i=1}^{n} (\tau_i - \tau_{i-1})^2, \\
r_{\text{dof\_acc}} &= -\lambda_{\text{dof\_acc}} \sum_{i=1}^{n} \left(\frac{\dot{q}_i - \dot{q}_{i-1}}{\Delta t}\right)^2.
\end{align}
where $\tau_i$ is the torque applied at the $i$-th time step, $\dot{q}_i$ is the joint velocity, and $\Delta t$ is the time step interval. These components penalize high torques, abrupt torque changes, and high joint accelerations.

\subsubsection{Collision Avoidance}
Avoiding collisions is critical for safe navigation within the pipe. The robot is penalized when it makes contact with obstacles or pipe walls, discouraging unsafe behavior. The collision avoidance reward is defined as\cite{rudin2022learning}:
\begin{equation}
r_{\text{collision}} = -\lambda_{\text{collision}} \sum_{i=1}^{n} \mathbb{I}(\|\mathbf{f}_{\text{contact}, i}\| > \text{threshold})
\end{equation}
where $\mathbf{f}_{\text{contact}, i}$ represents the contact forces on penalized contact indices, and the indicator function $\mathbb{I}(\cdot)$ signals a penalty when the force exceeds a specified threshold.

\subsubsection{Centerline Distance Penalty}
\begin{figure}
    \centering
    \includegraphics[width=0.8\linewidth]{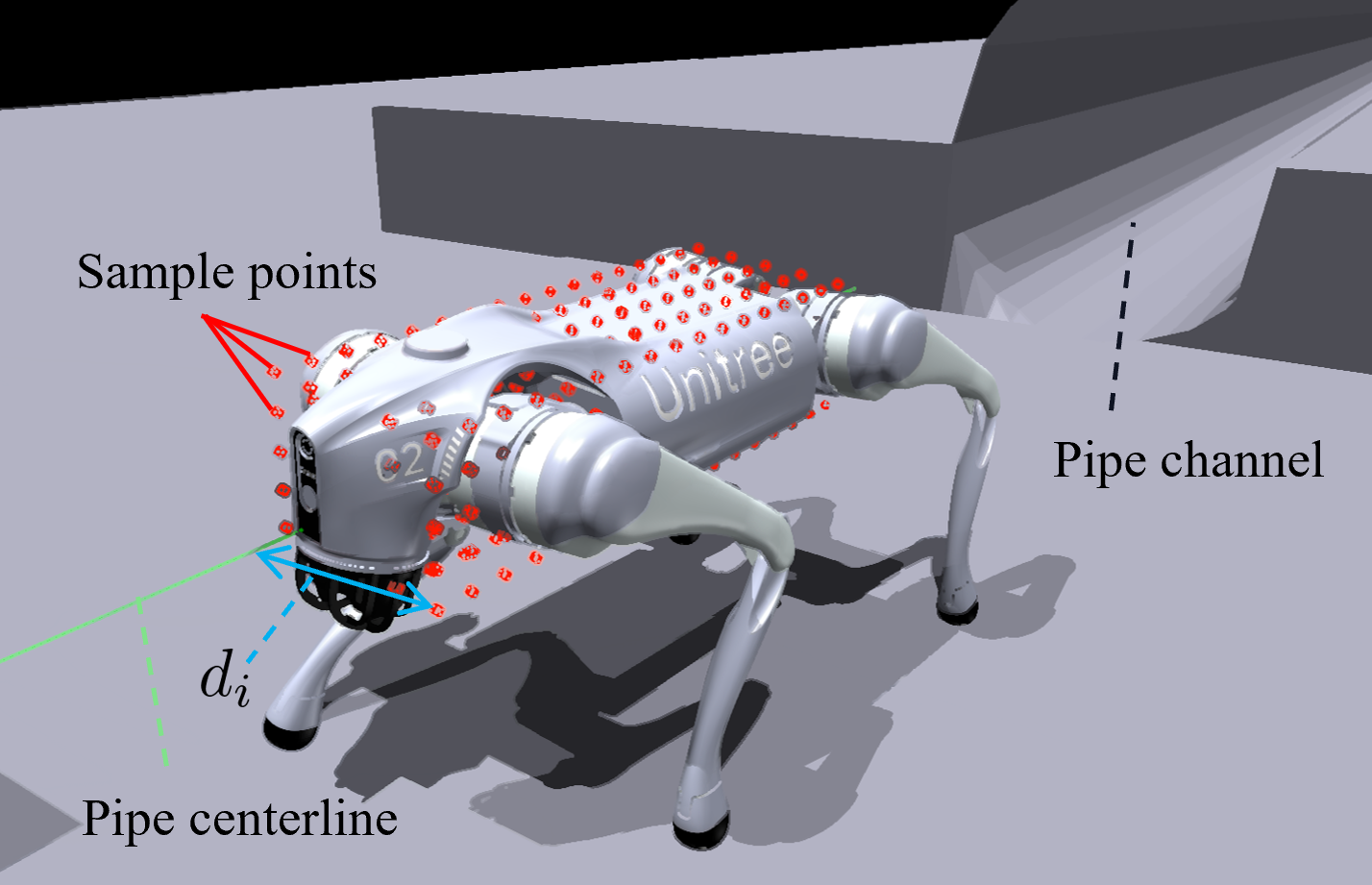}
    \caption{Centerline rewards. We sample those points on the robot base and compute the reward based on the distances between those points and the pipe centerline.}
    \label{fig:centerreward}
    \vspace{-6mm}
\end{figure}

To ensure the robot remains close to the center of the pipe, the centerline distance penalty is used. This reward component penalizes deviations from the centerline, thereby encouraging stable navigation within the confined environment. The centerline distance reward is expressed as:
\begin{equation}
r_{\text{centerline}} = -\lambda_{\text{centerline}} \left(\frac{1}{n} \sum_{i=1}^{n} d_i\right)
\end{equation}
As shown in Fig.~\ref{fig:centerreward} where $d_i$ are the distances from sampled points to the centerline of the pipe. This reward encourages the robot to stay near the pipe's central axis.

The scalar of the reward used in the training stage is comprehensively illustrated in Table~\ref{Rewardterm}.
\begin{table}[h]
    \centering
    \caption{Weights of the reward during training}
    \begin{tabular}{cc}
        \toprule
        Reward scalars & \makecell[c]{Value} \\ 
        \midrule
        $\lambda_{\text{lin\_vel}}$ & 1.5\\
        $\lambda_{\text{ang\_vel}}$ & 0.5\\
        $\lambda_{\text{torque}}$ & 1e-5\\
        $\lambda_{\text{delta\_torques}}$ & 1e-7\\
        $\lambda_{\text{dof\_acc}}$ & 2.5e-7\\
        $\lambda_{\text{collision}}$ & 10\\
        $\lambda_{\text{centerline}}$ & 0.3\\
        \bottomrule 
    \end{tabular}
    \vspace{-3mm}
    \label{Rewardterm}
\end{table}

\section{EXPERIMENTS}
\subsection{Simulation Setup}
We train our policy in the Isaac Gym simulation environment, leveraging the open-source framework Legged gym\cite{rudin2022learning}. The actor-critic algorithm utilizes Proximal Policy Optimization (PPO) with a clipping range of 0.2, generalized advantage estimation factor $\lambda$, and discount factor $\gamma$ are set to 0.95, and 0.99, respectively. The learning rate is set to 1e-3. To bridge the sim-to-real gap, we set some domain randomization when doing training. Specifically, we set the friction coefficient in the range of $[0.6,2]$ and the added base mass of $[0,3~\text{kg}]$.

\subsection{Training Process}
Training the robot to navigate through complex pipeline environments poses significant challenges due to the high difficulty of the task and the potential for sparse rewards when directly tackling difficult scenarios. To address this, the training process is divided into three progressive stages, each designed to incrementally build the robot’s capabilities and enhance the learning process.
\begin{itemize}
    \item \textbf{Basic gait:} In the first stage, the robot is trained in a relatively wide pipe with a larger diameter, allowing it to learn a stable and foundational gait without excessive constraints. This stage focuses on developing basic movement skills and familiarizing the policy with pipeline traversal.
    \item \textbf{Narrow pipe:} After establishing a basic gait, the second stage reduces the pipe diameter to present a more confined environment. This stage forces the robot to refine its movements and adapt its gait to the tighter space, improving its ability to maintain stability and maneuver effectively within narrower pipelines.
    \item \textbf{With obstacles:} In the final stage, various obstacles are introduced within the pipeline, including irregular protrusions and random obstructions. This phase is designed to challenge the robot’s adaptability and robustness, encouraging it to learn more complex avoidance strategies and further refine its movements to successfully navigate through unpredictable environments.
\end{itemize}

The training details in three stages are illustrated in Table~\ref{Stage}.
The training is conducted in parallel across 6144 robot agents on a PC equipped with a 32-core Intel i7- 13700 CPU, 64GB RAM, and an NVIDIA GEFORCE RTX 4090 GPU. The training process runs for around 15000 iterations in the first stage, 8000  iterations in the second stage and 8000 in the third stage. Taking approximately 10 hours in the first stage, 7 hours in the second and the third stage to complete.
\begin{table}[h]
    \centering
    \caption{Parameters in hierarchical training stages}
    \begin{tabular}{ccc}
        \toprule
        Training stage & \makecell[c]{Radius range} & \makecell[c]{Random obstacles} \\ 
        \midrule
        1 & $[0.3~\text{m},0.5 ~\text{m}]$ & w/o \\
        2 & $[0.2~\text{m},0.3~\text{m}]$ & w/o \\
        3 & $[0.2~\text{m},0.3~\text{m}]$ & w \\
        \bottomrule 
    \end{tabular}
    \label{Stage}
    \vspace{-4mm}
\end{table}

\begin{figure*}[t]
    \centering
    \includegraphics[width=1\linewidth]{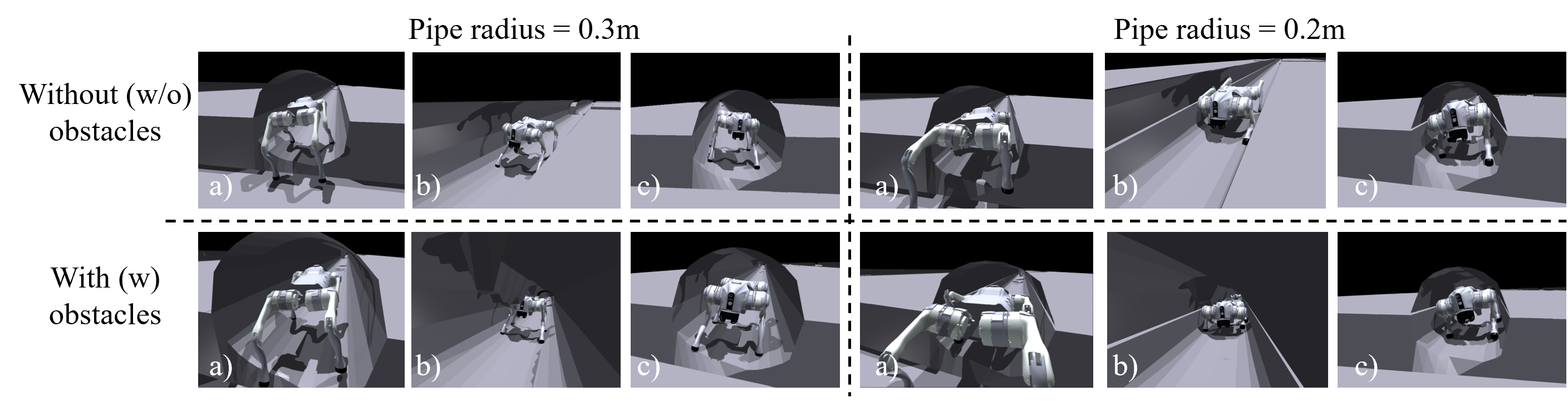}
    \vspace{-4mm}
    \caption{The screenshots of simulation tests of quadrupedal robot's pipe crossing tasks utilizing the trained policies obtained from the RL approach in 4 different cases. The upper and lower rows represent the pipe crossing without (w/o) and with (w) random obstacles, respectively. The left groups are within the pipes with a radius of 0.3 m, and the right groups are with a radius of 0.2 m. For each group, from a) to c), the quadrupedal robot is stepping into the entrance of the pipe, going through the pipe with the trained RL policies, and going out of the pipe. }
    \label{fig:pipeCross_sim}
\end{figure*}

\subsection{Simulation Results}
We evaluate the success rate and the average complete time to cross the pipe for the policies trained at different stages on various pipe terrains, as shown in Table ~\ref{evaluate}. In the first stage, the policy handles simpler conditions well, with high success rates and relatively quick traversal times in wide, unobstructed pipes. However, when faced with narrower pipes or the introduction of obstacles, the policy struggles, showing a significant decline in performance, characterized by lower success rates and increased traversal times.

As the training advances to the second stage, the policy becomes more adept at navigating narrower pipes, achieving consistent success across varying pipe diameters. The policy’s traversal time also improves, indicating a refined gait adapted to the challenging geometry. Nonetheless, the presence of obstacles remains a critical challenge, where the policy's ability to navigate complex environments is still limited.

By the final stage, the policy demonstrates robust performance across all tested conditions, successfully navigating through narrow pipes both with and without randomly generated obstacles. The improvement in success rate and reduction in traversal time compared to earlier stages highlight the effectiveness of the three-stage hierarchical training approach.

Among the above pipe crossing tests, the progresses of 4 selected different simulation cases of quadrupedal robot utilizing the trained policies obtained from the proposed RL approach are shown in Fig.~\ref{fig:pipeCross_sim}.

\begin{table}[h]
\centering
\caption{Performance of policies trained in different stages on pipe navigation tasks}
\setlength{\tabcolsep}{0.4mm}{
\begin{tabular}{ccccccc}
\hline
Training & Pipe &Pipe& Random & Test & Success & Completion \\
Stage &Radius& Length &Obstacles & Times & Rate &  Time(Avg.)\\
\hline
1 & 0.3m & 18m & w/o & 16 & 100\%   & 22.36s \\
1 & 0.2m & 18m & w/o & 16 & 0\%     & - \\
1 & 0.3m & 18m & w   & 16 & 43.75\% & 53.61s \\
2 & 0.3m & 18m & w/o & 16 & 100\%   & 18.96s \\
2 & 0.2m & 18m & w/o & 16 & 100\%   & 24.58s \\
2 & 0.2m & 18m & w   & 16 & 0\%     & -      \\
3 & 0.3m & 18m & w/o & 16 & 100\%   & 18.74s \\
3 & 0.2m & 18m & w/o & 16 & 100\%   & 22.74s \\
3 & 0.2m & 18m & w   & 16 & 93.75\% & 25.32s \\
\hline
\end{tabular}}
\label{evaluate}
\end{table}

\subsection{Experimental Results}

A quadrupedal robot Unitree Go2 with 12 joints is used for experiments. The width of the robot is 0.31 m and the standard standing height is 0.4 m. For exteroception, Go2 is equipped with Intel RealSense D435 on the head. We crop the depth image to 58*87 and clip the depth to [-0.5,0.5] in both simulation and physical experiment.

\begin{figure*}[!h]
    \centering
    \includegraphics[width=1\linewidth]{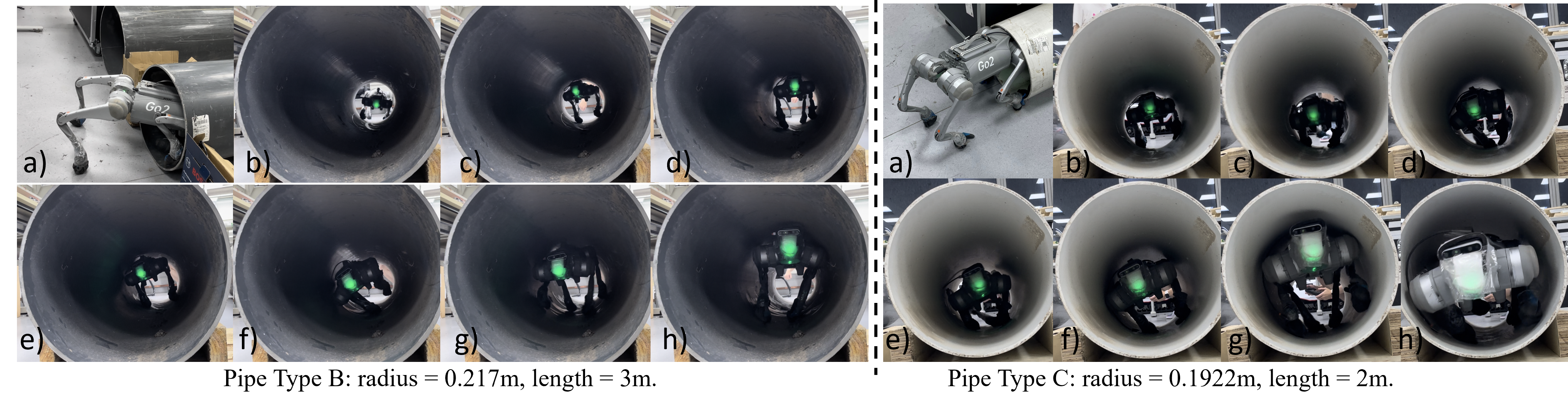}
    \vspace{-5mm}
    \caption{Experimental screenshots of the robot crossing the Pipe Type B and Pipe Type C without obstacles based on the proposed RL approach. For both: a) Stepping into the pipe entrance. b) to g) Crossing the pipe. h) Approaching the exit. }
    \label{fig:exp_cross}
    \vspace{-3mm}
\end{figure*}

To testify the trained policy of the proposed RL approach on real pipeline terrain, PVC pipes with different diameters were purchased and set up. To match the conditions of the simulation tests, the radius of the pipes are chosen from 0.19m to 0.24m, but only with 2 or 3 meters in length. The slippery PVC material, along with sand and dirt on its inner surface, made the experiments even more challenging. As a result, slipping happened frequently during the pipe-crossing tests.
Nevertheless, the quadrupedal robot can still adjust gaits and restore balance adaptively, and finally struggled to go through the pipe with the trained RL policies. The pipe-crossing processes of one of the successful experiments is shown in Fig.~\ref{fig:exp_cross}. After many trails, we can find that due to the inevitable sim-to-real gap, the success rate of the real experiments is much lower than that of simulation tests with the same trained policies. 
Various real-world physical factors, such as the swaying of unfixed pipes, the slippery inner surface, and the noisy onboard sensory information, would result in an unpredictable gap between simulation and real-world conditions. As a consequence, despite high success rates in simulation, the success rate in real-world experiments decreases.

The statistical results of the selected success real-world experiments can be seen in Table~\ref{tab_exp_results}. 
It can be concluded that the quadrupedal robot can achieve pipe-crossing tasks inside the 3 pipe type with a radius of 0.242 m, 0.217m, and 0.192m with an average speed of 0.176m/s, 0.150m/s, 0.111m/s. The quadrupedal robot can also deal with those circumstances which has obstacles inside.

\begin{table}[!h]
\centering
\caption{Experiment results}
\setlength{\tabcolsep}{0.3mm}{
\begin{tabular}{ccccccc}
\hline
\small{Pipe} & \small{Radius} & \small{Length} & \small{Random} & \small{Traversing} & \small{Completion} & \small{Crossing} \\ 
\small{Type} & \small{(m)}& \small{(m)} & \small{Obstacles}  & \small{Percentage} &  \small{Time(Avg.)} & \small{Speed(Avg.)} \\
\hline
A & 0.242 &3& w/o  &  100\% & 17s & 0.176m/s\\
B & 0.217 &3& w/o  &  100\% & 20s & 0.150m/s\\
C & 0.192 &2& w/o  &  100\% & 18s & 0.111m/s\\
C & 0.192 &2& w    &  46.5\% & 12s & 0.078m/s\\
\hline
\end{tabular}}
\label{tab_exp_results}
\vspace{-2mm}
\end{table}

\section{CONCLUSIONS}

In conclusion, we presented an efficient RL based learning method for training a policy to achieve narrow pipe crossing that is the core but challenging task in pipe inspection using quadrupedal robots. 
We firstly defined a new privileged visual information, i.e. bidirectional scandots, to obtain the terrain information of the pipe. Then, a new reward function that is suitable for quadrupedal robot navigation with narrow pipes was designed. 
We also tested our method in both simulation and real-world scenarios, both of which demonstrated the feasibility and adaptability of the proposed method. 
However, our current method may deteriorate when the visual input has a large noise or the robot is stuck by some unseen obstacles. 
In the future, a potential approach is to utilize LiDAR data as the onboard sensory information, which could be well-suited for pipe inspection scenarios.










\bibliographystyle{IEEEtran}
\normalem
\balance
\bibliography{IEEEabrv, main}

\end{document}